\documentclass[letterpaper,journal]{IEEEtran}
\usepackage{amsmath,amsfonts}
\usepackage{algorithmic}
\usepackage{algorithm}
\usepackage{array}
\usepackage[caption=false,font=normalsize,labelfont=sf,textfont=sf]{subfig}
\usepackage{textcomp}
\usepackage{stfloats}
\usepackage{url}
\usepackage{verbatim}
\usepackage{graphicx}
\usepackage{booktabs}
\usepackage{amssymb}
\usepackage{bm}
\usepackage{tikz}
\usetikzlibrary{arrows.meta,calc,positioning}
\usepackage{cite}
\usepackage{balance}
\usepackage[
colorlinks=true,
citecolor=blue,
linkcolor=blue,
urlcolor=blue
]{hyperref}
\def\BibTeX{{\rm B\kern-.05em{\sc i\kern-.025em b}\kern-.08em
		T\kern-.1667em\lower.7ex\hbox{E}\kern-.125emX}}
\begin{document}
\title{Large Language Model-Assisted Adaptive Heuristic Search for Agile Earth Observation Satellite Scheduling}
\author{He Wang, Yeye Liu, Ruijie Li, Xuemei Jiang, Junyu Wu, Lining Liu  \IEEEmembership{Member,~IEEE}, \\ Zhuoqi Man, Tao Li \IEEEmembership{Senior Member,~IEEE} and Yanjie Song \IEEEmembership{Senior Member,~IEEE}
	\thanks{This research was jointly funded by the Hainan Provincial Natural Science Foundation of 
		China (No.626QN0688),  National Natural Science Foundation of China (72501042, 62373117, 62403158, 62573150, 62503128),  Natural Science Foundation of Shandong Province under Grant (ZR2026QC0798) and Fundamental Research Funds for Central Universities (3072025GH0401). (Corresponding
		author: Junyu Wu and Lining Liu)
		
		He Wang, Junyu Wu, Lining Liu and Tao Li are with the College of Intelligent Science and Engineering, Harbin Engineering University, Harbin 150001, China (e-mail: wang\_he@hrbeu.edu.cn,  wujunyu@hrbeu.edu.cn, liningliu@163.com, taoli321@yahoo.com).
		
		Yeye Liu is with the School of Electrical and Control Engineering, North University of China, Taiyuan 030051, China (e-mail: 20240042@nuc.edu.cn).

		Ruijie Li is with the School of Electrical and Electronic Engineering, Shandong University of Technology, Zibo, 255000, China (liruijie@sdut.edu.cn).
		
		Xuemei Jiang is with the School of Unmanned Aerial Vehicle Industry, Chengdu Aeronautic Polytechnic University, Chengdu 610042, China
		(xuemeijane@126.com).
		
		Zhuoqi Man is with State Key Laboratory-Robot Research Institute, HIT, Harbin, Heilongjiang, 150001, China (e-mail: 23b308002@stu.hit.edu.cn).
		
		Yanjie Song is with the School of Information Science and Technology, Dalian Maritime University, Dalian 116026, China (e-mail: songyj\_2017@163.com).		
	}
}

\markboth{Journal of \LaTeX\ Class Files,~Vol.~18, No.~9, September~2020}%
{How to Use the IEEEtran \LaTeX \ Templates}

\maketitle

\begin{abstract}
Agile Earth observation satellites must coordinate imaging tasks within limited observation windows, while accounting for attitude maneuvering and onboard energy constraints. Competition among observation opportunities makes scheduling performance sensitive to the heuristic rules used for task selection. This letter proposes a large language model (LLM)-assisted adaptive heuristic search method for agile Earth observation satellite scheduling. The LLM generates an initial heuristic rule according to the characteristics of the scheduling instance. During the subsequent ant colony search, search performance and scheduling information are fed back to the LLM to refine the heuristic rule. The proposed modifications are evaluated through paired solution constructions before being incorporated into the search process. A unified scheduling procedure determines task selection, satellite assignment, and observation timing under the operational constraints. Experiments on six scheduling scenarios involving 50--200 imaging tasks demonstrate that the proposed method achieves 4.3\%--9.6\% higher mean observation benefit than the strongest baseline.
\end{abstract}

\begin{IEEEkeywords}
Remote Sensing, Earth Observation, Satellite Scheduling, Large Language Model, Heuristic Algorithm.
\end{IEEEkeywords}

\section{Introduction}

\IEEEPARstart{A}{gile} Earth observation satellites (AEOSs) can adjust their attitudes to image multiple ground targets within limited visibility intervals. However, scheduling these observations requires coordinating task selection, satellite assignment, and observation timing under attitude maneuvering and onboard resource constraints. As imaging requests increase, competition for observation windows becomes more pronounced, and executing one task may reduce the opportunities available to others. Efficient scheduling is essential for improving observation benefit under limited satellite resources~\cite{ref1}.

Considerable research has addressed AEOS scheduling through mathematical optimization and heuristic methods. Peng et al.~\cite{ref2} developed a branch-and-cut-and-price algorithm considering time-dependent transitions and variable observation windows. Rocha et al.~\cite{ref3} investigated integrated observation and downlink scheduling using mixed-integer programming and model-based heuristics. Wang et al.~\cite{ref4} proposed a Lagrangian relaxation-based heuristic to address time-dependent scheduling constraints. Although these methods provide effective optimization strategies, the large number of task sequences and satellite-window combinations makes efficient schedule construction an important research direction.

Heuristic and learning-assisted methods have also received increasing attention. Chen et al.~\cite{ref5} combined differential evolution with ensemble heuristics for satellite scheduling, while Li et al.~\cite{ref6} integrated deep Q-learning with heuristic task allocation. Chen et al.~\cite{ref7} developed a neural priority model to guide observation sequencing. More recently, Huang et al.~\cite{ref8} introduced reinforcement learning into dung beetle optimization to regulate search strategies. Wang et al.~\cite{ref9} employed reinforcement learning to guide evolutionary policy optimization, and Wei et al.~\cite{ref10} used genetic programming to evolve scheduling heuristics through knowledge transfer. These studies have advanced adaptive search and heuristic construction for AEOS scheduling. In particular, candidate evaluation remains important because task benefits, observation-window availability, and resource consumption may influence scheduling decisions differently across instances.

Large language models (LLMs) offer a way to generate and refine heuristic rules using optimization feedback. Romera-Paredes et al.~\cite{ref11} demonstrated that LLM-generated priority functions can be improved through external performance evaluation. Van Stein and B\"ack~\cite{ref12} further developed an LLM-based framework for automatically generating and refining metaheuristic algorithms. These approaches provide a basis for incorporating LLM-generated heuristics into AEOS scheduling. Rather than directly producing observation sequences, an LLM can construct task-selection rules from instance information and revise them according to search performance, while the scheduling algorithm handles feasible solution construction.

This letter proposes a large language model-assisted adaptive heuristic search method for AEOS scheduling. An LLM generates an initial candidate-observation scoring rule and subsequently refines it using feedback from ant colony search. The proposed modifications are evaluated before being incorporated into subsequent search. The main contributions are summarized as follows:

\begin{enumerate}
\item An LLM-assisted initialization method is developed to generate and validate instance-specific scheduling heuristics.
\item A feedback-guided adaptation mechanism is introduced to refine heuristics during ant colony search through bounded updates and paired evaluations.
\item A unified scheduling framework integrates
LLM-generated candidate scoring with
constraint-preserving solution construction,
enabling effective search across different
task scales and observation-window conditions.	
\end{enumerate}

\section{Scheduling Model and LLM-Assisted Search}
\label{sec:model}

\subsection{Scheduling Model}

Consider a set of imaging tasks
$\mathcal{I}=\{1,\ldots,N_I\}$ and a set of agile
Earth observation satellites
$\mathcal{S}=\{1,\ldots,N_S\}$ over a planning
horizon $\mathcal{T}=[0,T]$.
Each task $i\in\mathcal{I}$ is characterized by
its priority $p_i$, imaging duration $d_i$,
release time $r_i$, and deadline $D_i$.
For satellite $s\in\mathcal{S}$, let
$\mathcal{W}_{is}$ denote the set of available
observation windows for task $i$.
Each window indexed by
$k\in\mathcal{W}_{is}$ is characterized
by its visibility interval $[a_{isk},b_{isk}]$,
observation quality $q_{isk}$, and required
pointing angle $\theta_{isk}$.

A candidate observation is represented by
$z=(i,s,k)$, and all candidate observations
form the set
\begin{equation}
	\mathcal{Z}=
	\left\{(i,s,k)\mid
	i\in\mathcal{I},\,
	s\in\mathcal{S},\,
	k\in\mathcal{W}_{is}
	\right\}.
	\label{eq:candidate_set}
\end{equation}
Let $x_{isk}\in\{0,1\}$ indicate whether
candidate $(i,s,k)$ is scheduled.
The objective is to maximize the total
observation benefit:
\begin{equation}
	\max F=
	\sum_{i\in\mathcal{I}}
	\sum_{s\in\mathcal{S}}
	\sum_{k\in\mathcal{W}_{is}}
	p_iq_{isk}x_{isk}.
	\label{eq:objective}
\end{equation}
Each imaging task can be executed at most once:
\begin{equation}
	\sum_{s\in\mathcal{S}}
	\sum_{k\in\mathcal{W}_{is}}
	x_{isk}\leq 1,
	\quad \forall i\in\mathcal{I}.
	\label{eq:task_unique}
\end{equation}

The complete schedule is denoted by
$\mathcal{P}=\{\Pi_s\mid s\in\mathcal{S}\}$,
where
$\Pi_s=(z_{s1},\ldots,z_{sm_s})$
is the chronologically ordered observation
sequence assigned to satellite $s$.
For a selected candidate $z=(i,s,k)$,
its imaging start time $t_z$ must satisfy
\begin{equation}
	\begin{aligned}
		\underline{t}_z
		&=\max\{r_i,a_{isk}\},\\
		\overline{t}_z
		&=\min\{D_i,b_{isk}\}-d_i,\\
		\underline{t}_z
		&\leq t_z\leq\overline{t}_z.
	\end{aligned}
	\label{eq:observation_window}
\end{equation}
The observation sequence must also satisfy
the attitude-transition and onboard
energy constraints described below.

For two consecutive observations
$u=z_{s\ell}$ and $v=z_{s,\ell+1}$
in $\Pi_s$, let $\Delta\theta_{uv}$
denote the required attitude displacement.
The corresponding maneuver time is
determined by the maximum angular
velocity $\omega_s^{\max}$ and angular
acceleration $a_s^{\max}$:
\begin{equation}
	T_s(\Delta\theta)=
	\begin{cases}
		2\sqrt{\dfrac{\Delta\theta}{a_s^{\max}}},
		& \Delta\theta\leq
		\dfrac{(\omega_s^{\max})^2}{a_s^{\max}},\\[6pt]
		\dfrac{\Delta\theta}{\omega_s^{\max}}
		+\dfrac{\omega_s^{\max}}{a_s^{\max}},
		& \text{otherwise}.
	\end{cases}
	\label{eq:slew_time}
\end{equation}
For simplicity, let
$T_{uv}=T_s(\Delta\theta_{uv})$.
The start times of adjacent observations
must satisfy
\begin{equation}
	t_v\geq t_u+d_{i(u)}+T_{uv},
	\quad \forall (u,v)\in\mathcal{A}_s,
	\label{eq:transition_constraint}
\end{equation}
where $\mathcal{A}_s$ denotes the set
of consecutive observation pairs in
$\Pi_s$, and $i(u)$ identifies the task
associated with observation $u$.

Let $P_s^{\mathrm{img}}$ and
$P_s^{\mathrm{man}}$ denote the imaging
and maneuvering powers of satellite $s$,
respectively. Its total energy consumption
is expressed as
\begin{equation}
	E_s(\Pi_s)=
	\frac{1}{\kappa_E}
	\left[
	P_s^{\mathrm{img}}
	\sum_{\ell=1}^{m_s}d_{i(z_{s\ell})}
	+
	P_s^{\mathrm{man}}
	\sum_{\ell=1}^{m_s}
	T_s(\Delta\theta_{s\ell})
	\right],
	\label{eq:energy_consumption}
\end{equation}
where $\Delta\theta_{s\ell}$ is the
attitude displacement from the preceding
observation, or from the initial attitude
for the first observation, and
$\kappa_E$ is the energy unit conversion
factor. The energy constraint is
\begin{equation}
	E_s(\Pi_s)\leq E_s^{\max},
	\quad \forall s\in\mathcal{S},
	\label{eq:energy_constraint}
\end{equation}
where $E_s^{\max}$ is the available
onboard energy. Together with
(\ref{eq:task_unique}) and
(\ref{eq:observation_window}), these
constraints define the feasible
scheduling space.

\subsection{Feasible Observation Construction}

A feasible schedule is constructed by inserting
candidate observations into satellite-specific
sequences. Given the current schedule
$\mathcal{P}=\{\Pi_s\mid s\in\mathcal{S}\}$,
let $\mathcal{G}_s(\Pi_s)$ denote the set of
insertion positions in the sequence of
satellite $s$. For a candidate
$z=(i,s,k)$, each position is determined
by its preceding observation $u$ and
succeeding observation $v$.

The admissible start-time interval of
candidate $z$ at position $(u,v)$ is
\begin{equation}
	\begin{aligned}
		\underline{t}_z^{(u,v)}
		=&\max\bigl\{
		r_i,a_{isk},
		t_u+d_{i(u)}+T_{uz}
		\bigr\},\\
		\overline{t}_z^{(u,v)}
		=&\min\bigl\{
		D_i-d_i,b_{isk}-d_i,
		t_v-d_i-T_{zv}
		\bigr\}.
	\end{aligned}
	\label{eq:insertion_interval}
\end{equation}
Here, $T_{uz}$ and $T_{zv}$ denote the
maneuver times associated with the two
new transitions. The initial satellite
state and planning horizon provide the
boundary conditions when $u$ or $v$
does not exist. A candidate can be
inserted without shifting the existing
observations if
\begin{equation}
	\underline{t}_z^{(u,v)}
	\leq
	\overline{t}_z^{(u,v)}.
	\label{eq:insertion_feasibility}
\end{equation}

Inserting $z$ also changes the maneuvering
energy of the corresponding satellite.
For an internal insertion position,
the incremental energy consumption is
\begin{equation}
	\begin{aligned}
		\Delta E_s(z\mid u,v)
		=&\frac{P_s^{\mathrm{img}}d_i}{\kappa_E}\\
		&+\frac{P_s^{\mathrm{man}}}{\kappa_E}
		\left(T_{uz}+T_{zv}-T_{uv}\right).
	\end{aligned}
	\label{eq:incremental_energy}
\end{equation}
The last term removes the original
transition from $u$ to $v$. At sequence
boundaries, the corresponding initial
attitude or terminal condition is used.
The insertion is accepted only when
\begin{equation}
	E_s(\Pi_s)+\Delta E_s(z\mid u,v)
	\leq E_s^{\max}.
	\label{eq:insertion_energy}
\end{equation}

\begin{figure}[!t]
	\centering
	\begin{tikzpicture}[
		x=0.82cm,
		y=0.73cm,
		font=\scriptsize,
		>=Latex,
		obs/.style={
			draw=black!75,
			fill=blue!18,
			rounded corners=1.5pt,
			line width=0.5pt
		},
		candidate/.style={
			draw=blue!70!black,
			fill=blue!40,
			rounded corners=1.5pt,
			line width=0.7pt
		},
		alternative/.style={
			draw=blue!65!black,
			fill=white,
			dashed,
			rounded corners=1.5pt,
			line width=0.7pt
		},
		window/.style={
			draw=blue!50,
			fill=blue!7,
			dashed,
			rounded corners=2pt
		}
		]
		
		\node[anchor=west] at (0.1,3.30)
		{\textbf{Satellite $s$}};
		
		\node[anchor=west] at (0.1,1.25)
		{\textbf{Satellite $s'$}};
		
		\draw[->,gray!75] (0.2,2.20)--(9.25,2.20)
		node[right] {$t$};
		\draw[->,gray!75] (0.2,0.28)--(9.25,0.28)
		node[right] {$t$};
		
		\draw[window] (3.15,1.93)
		rectangle (5.15,2.83);
		
		\node[blue!65!black,font=\tiny]
		at (4.15,3.02) {Available window};
		
		\node[obs,minimum width=1.02cm,
		minimum height=0.43cm]
		(u) at (1.50,2.37) {$u$};
		
		\node[candidate,minimum width=1.02cm,
		minimum height=0.43cm]
		(z) at (4.12,2.37) {$z$};
		
		\node[obs,minimum width=1.02cm,
		minimum height=0.43cm]
		(v) at (7.13,2.37) {$v$};
		
		\draw[<->,red!70!black,thin]
		(2.16,2.58)--(3.45,2.58);
		\node[red!70!black,font=\tiny]
		at (2.81,2.85) {Maneuver};
		
		\draw[<->,red!70!black,thin]
		(4.80,2.58)--(6.44,2.58);
		\node[red!70!black,font=\tiny]
		at (5.62,2.85) {Maneuver};
		
		\draw[blue!45,densely dashed]
		(3.15,1.85)--(3.15,2.95);
		\draw[blue!45,densely dashed]
		(5.15,1.85)--(5.15,2.95);
		
		\draw[window] (3.15,0.42)
		rectangle (6.00,0.99);
		
		\node[alternative,
		minimum width=1.05cm,
		minimum height=0.38cm]
		at (4.30,0.70) {$z'$};
		
		\node[font=\tiny,anchor=west]
		at (6.15,0.71)
		{Alternative window};
		
		\draw[dashed,->,blue!65!black]
		(4.15,1.90)--(4.15,1.05);
		
		\node[font=\tiny,gray!80]
		at (1.50,1.83) {Scheduled};
		
		\node[font=\tiny,blue!70!black]
		at (4.12,1.83) {Insertion};
		
		\node[font=\tiny,gray!80]
		at (7.13,1.83) {Scheduled};
		
	\end{tikzpicture}
	\caption{Feasible observation insertion and
		alternative opportunities.}
	\label{fig:insertion}
\end{figure}
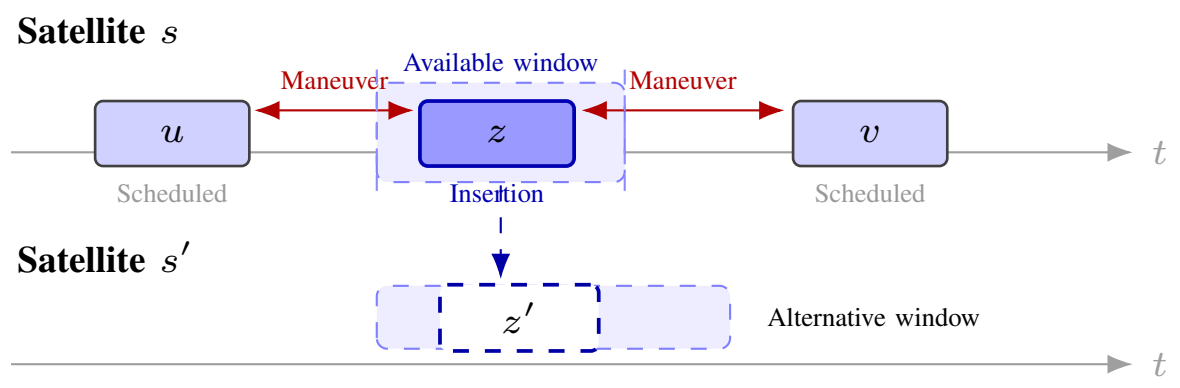

The insertion process is illustrated in
Fig.~\ref{fig:insertion}, where a candidate
observation is placed between two scheduled
tasks while accounting for maneuvering times
and alternative opportunities.

Let $\mathcal{F}$ denote the set of
schedules satisfying the constraints
in Section II-A. The feasible candidate
set under the current schedule is
\begin{equation}
	\mathcal{C}(\mathcal{P})=
	\left\{
	z\in\mathcal{Z}\ \middle|\
	\begin{array}{l}
		\exists\,\ell\in\mathcal{G}_s(\Pi_s),\\
		\mathcal{P}\oplus(z,\ell)\in\mathcal{F}
	\end{array}
	\right\},
	\label{eq:feasible_candidates}
\end{equation}
where $\mathcal{P}\oplus(z,\ell)$ denotes
the schedule obtained by inserting
candidate $z$ at position $\ell$.
The insertion decoder examines feasible
positions and determines the earliest
admissible execution time for each
candidate observation. The resulting
feasible candidates form
$\mathcal{C}(\mathcal{P})$.

\subsection{LLM-Assisted Heuristic Initialization}

At the beginning of each scheduling instance,
the LLM receives a summary of task benefits,
observation-window distributions, and satellite
resource conditions. Based on this information,
the LLM generates an initial heuristic scoring
function $g_0$ for evaluating candidate
observations. Let $\bm{f}_z$ denote the
normalized scheduling features of candidate
$z\in\mathcal{C}(\mathcal{P})$. The scoring
function consists of linear terms and
selected pairwise interactions:

\begin{equation}
	g_0(\bm{f}_z)=
	\sum_{a\in\mathcal{K}}w_af_{z,a}
	+\sum_{(a,b)\in\mathcal{J}}v_{ab}f_{z,a}f_{z,b}.
	\label{eq:initial_heuristic}
\end{equation}
where $\mathcal{K}$ denotes the feature
index set, $\mathcal{J}$ contains the
selected feature pairs, and $w_a$ and
$v_{ab}$ are the generated coefficients.

The LLM returns the scoring function
in a structured format. Feature names,
coefficient ranges, and numerical values
are validated before the function is
used in scheduling. The resulting heuristic
value is calculated as
\begin{equation}
	\eta_0(z)=
	\exp\left(
	\operatorname{clip}
	\left[g_0(\bm{f}_z),-C_g,C_g\right]
	\right),
	\label{eq:heuristic_value}
\end{equation}
where $C_g$ bounds the scoring function
and $\eta_0(z)>0$.

During schedule construction, the heuristic
is combined with pheromone information
to determine the selection probability
of each feasible candidate:
\begin{equation}
	\Pr_t(z\mid c,\mathcal{P})=
	\frac{\tau_{cz}^{\alpha}
		\eta_t(z)^{\beta}}
	{\displaystyle\sum_{\xi\in\mathcal{C}(\mathcal{P})}
		\tau_{c\xi}^{\alpha}\eta_t(\xi)^{\beta}},
	\label{eq:selection_probability}
\end{equation}
where $c$ denotes the preceding candidate
in the construction order, $\tau_{cz}$
is the corresponding pheromone intensity,
and $\alpha$ and $\beta$ control the
relative influence of pheromone and
heuristic information. Initially,
$\eta_t(z)=\eta_0(z)$. The selected
candidate is inserted into the satellite
schedule using the procedure in
Section II-B.

After each search iteration, pheromone
intensities are updated according to
\begin{equation}
	\begin{aligned}
		\tau_{cz}\leftarrow
		\operatorname{clip}\bigl(
		&(1-\rho)\tau_{cz}+\Delta\tau_{cz},\\
		&\tau_{\min},\tau_{\max}
		\bigr),
	\end{aligned}
	\label{eq:pheromone_update}
\end{equation}
where $\rho$ is the evaporation rate,
$\Delta\tau_{cz}$ denotes the pheromone
reinforcement from the selected search
solution, and $\tau_{\min}$ and
$\tau_{\max}$ are the pheromone bounds.
The heuristic rule is subsequently
adjusted using search feedback.

\subsection{Feedback-Based Heuristic Adaptation}

During ant colony search, feedback is collected
at two checkpoints corresponding to 20\% and
45\% of the evaluation budget. Let
$\mathcal{H}_t$ denote the feedback information
at checkpoint $t$, including the best
observation benefit, search stagnation,
unassigned high-priority tasks, satellite
energy utilization, and pheromone statistics.
The LLM uses $\mathcal{H}_t$ and the current
heuristic rule to generate a residual
scoring function $h_t$.

The updated heuristic score is defined as

\begin{equation}
	g_t(\bm{f}_z)=\operatorname{clip}
	\left(g_0(\bm{f}_z)+\widehat{h}_t(\bm{f}_z),-C_g,C_g\right).
	\label{eq:adaptive_score}
\end{equation}
where
$\widehat{h}_t(\bm{f}_z)=
\operatorname{clip}
(h_t(\bm{f}_z),-C_h,C_h)$
is the bounded residual and $C_h$ controls
the maximum adjustment. The corresponding
heuristic value is
\begin{equation}
	\eta_t(z)=\exp\left(g_t(\bm{f}_z)\right).
	\label{eq:adaptive_heuristic}
\end{equation}
Each proposed residual is structurally
and numerically validated before evaluation.

To assess a proposed modification,
the current and updated rules are
evaluated through $M$ paired schedule
constructions using identical pheromone
states and random seeds. The mean
benefit difference is calculated as
\begin{equation}
	\Delta F_t=
	\frac{1}{M}
	\sum_{j=1}^{M}
	\left(F_j^{\mathrm{new}}-
	F_j^{\mathrm{cur}}\right),
	\label{eq:feedback_gain}
\end{equation}
where $F_j^{\mathrm{new}}$ and
$F_j^{\mathrm{cur}}$ denote the observation
benefits obtained using the proposed
and current heuristics, respectively.

The proposed rule is accepted when
$\Delta F_t>0$ and the number of
improved pairs exceeds that of
degraded pairs. The accepted residual
replaces the previous correction,
while the initial rule $g_0$ remains
unchanged. Subsequent search uses the
updated heuristic together with
pheromone information to construct
feasible schedules.

\begin{figure*}[!t]
	\centering
	\includegraphics[width=0.999\textwidth]{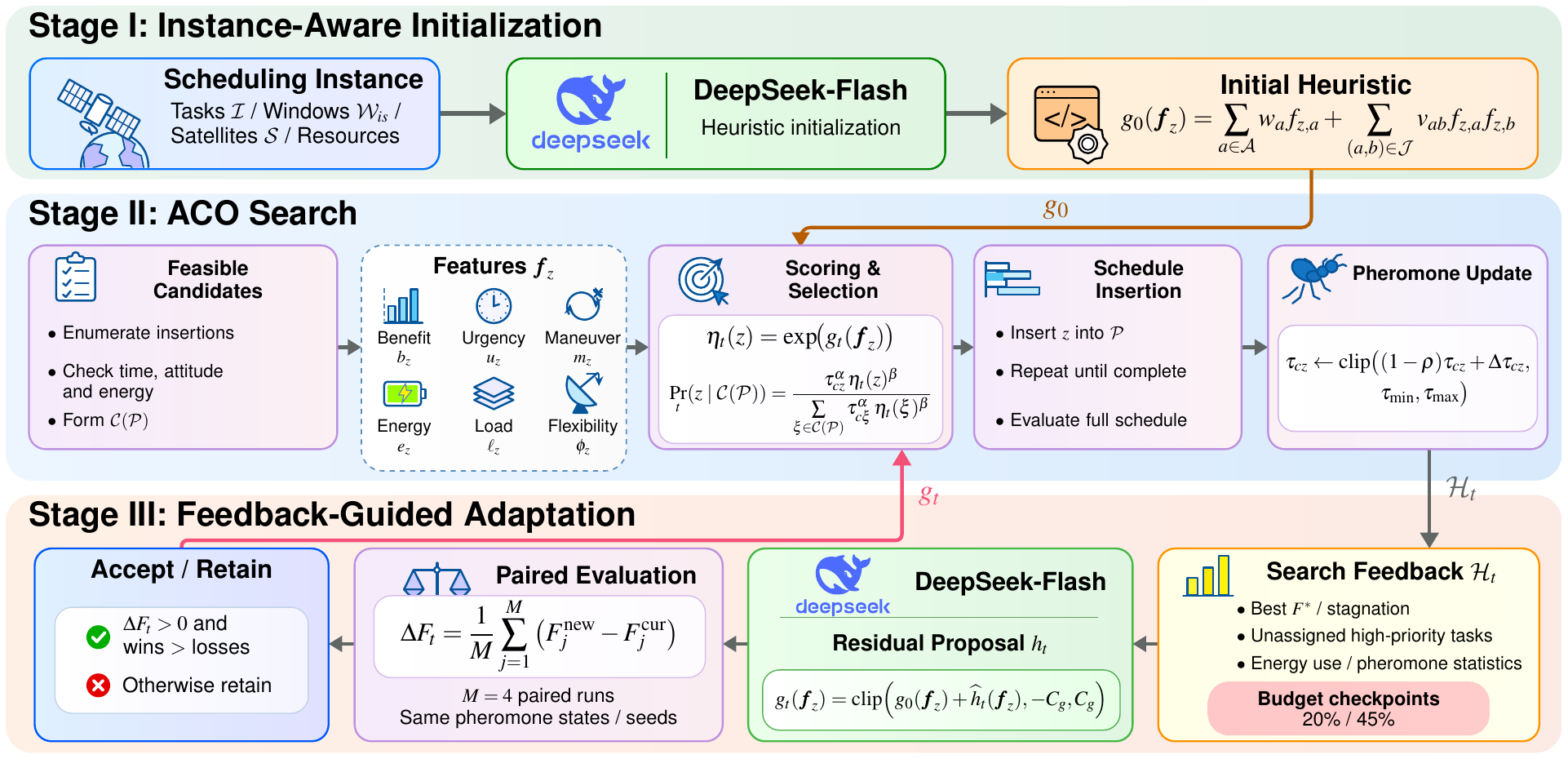}
	\caption{Framework of the proposed LLM-assisted adaptive heuristic search for AEOSs scheduling.}
	\label{fig:method}
\end{figure*}

The overall search procedure is illustrated
in Fig.~\ref{fig:method} and summarized in
Algorithm~\ref{alg:llmaco}.

\begin{algorithm}[!t]
	\caption{LLM-Assisted Adaptive Heuristic Search}
	\label{alg:llmaco}
	\begin{algorithmic}[1]
		\REQUIRE Scheduling instance $\mathcal{D}$,
		evaluation budget $B$
		\ENSURE Best feasible schedule $\mathcal{P}^{\star}$
		
		\STATE Extract instance information
		from $\mathcal{D}$
		\STATE Generate and validate initial
		LLM heuristic $g_0$
		\STATE Initialize pheromone $\tau$,
		residual $h=0$, and $\mathcal{P}^{\star}$
		
		\WHILE{search budget remains}
		
		\STATE Construct feasible schedules
		using (\ref{eq:selection_probability})
		
		\STATE Update $\mathcal{P}^{\star}$
		and pheromone by
		(\ref{eq:pheromone_update})
		
		\IF{a feedback checkpoint is reached}
		
		\STATE Collect search feedback
		$\mathcal{H}_t$
		
		\STATE Generate and validate
		residual proposal $\widehat{h}_t$
		
		\STATE Evaluate current and
		proposed rules using paired
		constructions
		
		\IF{$\Delta F_t>0$ and
			wins $>$ losses}
		
		\STATE $h\gets h_t$
		
		\ENDIF
		
		\ENDIF
		
		\ENDWHILE
		
		\STATE \textbf{return}
		$\mathcal{P}^{\star}$
		
	\end{algorithmic}
\end{algorithm}

\section{Experimental Results and Analysis}

\subsection{Simulation Setup}

Six scheduling scenarios, denoted as S1--S6,
are considered to evaluate the proposed method.
The task numbers are set to 50, 100, and 200,
with two scenarios at each scale.
S1, S3, and S5 represent moderate observation-window
competition, while S2, S4, and S6 represent
high competition. Each scenario involves three
agile satellites over a 24-h planning horizon.
The observation duration is 60 s, and task
priorities are randomly assigned from 1 to 10.
The scheduling process considers observation
windows, attitude maneuvering, and onboard
energy constraints.

Five representative algorithms are selected for
comparison, including ant colony optimization
(ACO)~\cite{ref13}, artificial bee colony
(ABC)~\cite{ref14}, genetic algorithm
(GA)~\cite{ref15}, particle swarm optimization
(PSO)~\cite{ref16}, and whale optimization
algorithm (WOA)~\cite{ref17}.
Their parameters follow the corresponding
reference settings.

The proposed LLMACO employs DeepSeek-Flash
for heuristic generation and adaptation,
with $\alpha=1$, $\beta=3$, and $\rho=0.15$
for ant colony search.
Heuristic updates are considered at 20\%
and 45\% of the maximum iteration count,
with four paired trials used to evaluate
each proposed update.
All algorithms are independently executed
20 times per scenario, with a maximum of
100 iterations per run. The mean and
standard deviation are reported for comparison.

\subsection{Scheduling Performance and Convergence}

\begin{figure}[!h]
	\centering
	\includegraphics[width=\columnwidth]
	{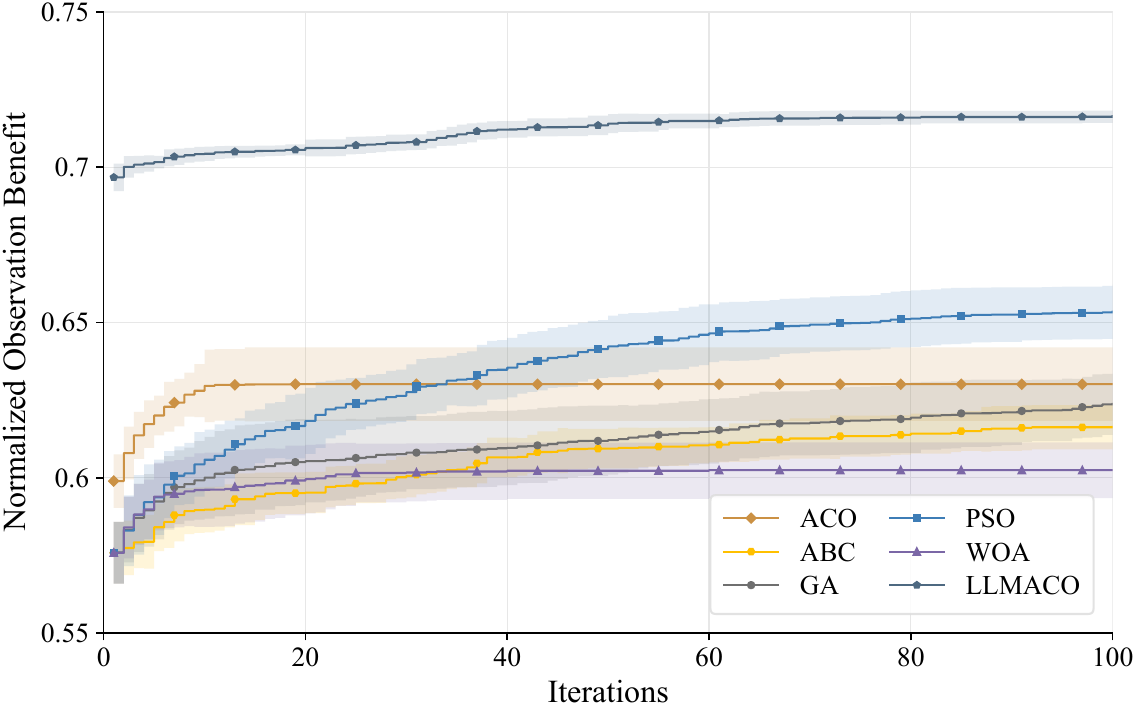}
	\caption{Convergence comparison of different
		algorithms under scenario S6.}
	\label{fig:convergence}
\end{figure}

Table~\ref{tab:results} and
Fig.~\ref{fig:convergence} present the
scheduling performance and convergence
behavior of the compared algorithms.
LLM-ACO achieves the highest mean
observation benefit in all six scenarios,
improving upon the strongest baseline
by 4.32\%--9.63\%. At each task scale,
the improvement is greater under high
observation-window competition.
LLM-ACO also exhibits the lowest
standard deviation across all scenarios.
In S6, the proposed method maintains
a clear advantage from the early
iterations, while PSO improves
progressively and conventional ACO
stabilizes earlier. These results
indicate that the generated heuristic
provides effective candidate-selection
preferences throughout the search.

\begin{table*}[!t]

\caption{Comparison of observation benefit
	after 100 iterations (mean $\pm$ standard
	deviation over 20 independent runs).
	The last column reports the improvement
	over the strongest baseline. Best values
	are highlighted in bold.}
	\label{tab:results}
	\centering
	\setlength{\tabcolsep}{4pt}\renewcommand{\arraystretch}{1.13}
	\scriptsize
	\begin{tabular*}{\textwidth}{@{\extracolsep{\fill}}l c c c c c c c@{}}
		\toprule
		Scenario & ACO & ABC & GA & PSO & WOA & LLMACO & Gain vs.\ best (\%)\\
		\midrule
		S1 & $225.70\pm4.82$ & $228.10\pm3.51$ & $227.30\pm5.06$ & $233.55\pm3.17$ & $221.05\pm5.24$ & $\mathbf{243.65\pm0.81}$ & $+4.32^{+}$\\
		S2 & $181.90\pm3.52$ & $191.05\pm2.82$ & $184.40\pm7.14$ & $191.85\pm4.39$ & $175.75\pm6.34$ & $\mathbf{202.50\pm0.51}$ & $+5.55^{+}$\\
		S3 & $489.80\pm6.01$ & $486.00\pm5.89$ & $489.60\pm9.91$ & $503.75\pm8.75$ & $476.10\pm7.85$ & $\mathbf{528.20\pm1.58}$ & $+4.85^{+}$\\
		S4 & $416.35\pm5.59$ & $416.05\pm5.83$ & $419.05\pm8.99$ & $431.10\pm9.87$ & $397.80\pm8.91$ & $\mathbf{470.35\pm0.75}$ & $+9.10^{+}$\\
		S5 & $844.05\pm7.27$ & $817.90\pm9.16$ & $835.60\pm11.84$ & $866.05\pm12.19$ & $812.15\pm12.61$ & $\mathbf{920.55\pm2.26}$ & $+6.29^{+}$\\
		S6 & $690.60\pm13.29$ & $675.50\pm7.98$ & $683.65\pm10.93$ & $716.20\pm9.88$ & $660.25\pm10.08$ & $\mathbf{785.20\pm2.21}$ & $+9.63^{+}$\\
		\bottomrule
	\end{tabular*}
\end{table*}

\subsection{Computational Time Analysis}

\begin{figure*}[!t]
	\centering
	\includegraphics[width=0.999\textwidth]
	{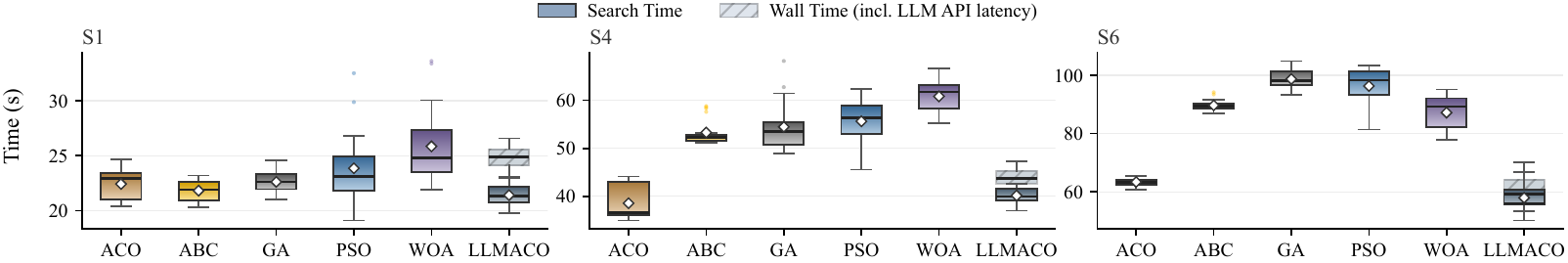}
	\caption{Runtime distributions of different
		algorithms under scenarios S1, S4, and S6.}
	\label{fig:runtime}
\end{figure*}

Fig.~\ref{fig:runtime} compares the runtime
distributions of the six algorithms in
S1, S4, and S6. Compared with ACO,
LLMACO requires additional processing
for heuristic generation and feedback
evaluation. Its runtime is slightly higher
in S1 and S4, but lower in S6.
The generated rules are evaluated directly
during schedule construction, while LLM
inference is restricted to predefined
feedback checkpoints. The results show
that adaptive heuristic search maintains
competitive computational efficiency
as the scheduling scale increases.

\section{Conclusion}

This letter investigated LLM-assisted
heuristic search for agile Earth observation
satellite scheduling. The proposed LLMACO
achieved consistently higher observation
benefits across different task scales and
observation-window competition levels.
The results indicate that instance-specific
heuristic rules can improve candidate
selection while preserving the feasibility
of constructed schedules. The current
experiments consider fixed task sets and
observation windows. Future research will
address dynamic task arrivals and changing
observation conditions, together with
the transfer of effective heuristic rules
between scheduling instances.


\begin{thebibliography}{99}
	
	\bibitem{ref1}
	B. Ferrari, J.-F. Cordeau, M. Delorme,
	M. Iori, and R. Orosei,
	``Satellite scheduling problems: A survey of
	applications in Earth and outer space observation,''
	\textit{Computers \& Operations Research},
	vol. 173, Art. no. 106875, 2025.
	\href{https://doi.org/10.1016/j.cor.2024.106875}
	{doi: 10.1016/j.cor.2024.106875}.
	
	\bibitem{ref2}
	G. Peng, J. Wang, G. Song, A. Gunawan,
	L. Xing, and P. Vansteenwegen,
	``Branch-and-cut-and-price for agile Earth
	observation satellite scheduling,''
	\textit{European Journal of Operational Research},
	vol. 326, no. 3, pp. 427--438, 2025.
	\href{https://doi.org/10.1016/j.ejor.2025.04.014}
	{doi: 10.1016/j.ejor.2025.04.014}.
	
	\bibitem{ref3}
	Y. Rocha, G. O. Chagas, L. C. Coelho,
	and A. Subramanian,
	``The integrated agile Earth observation
	satellite scheduling problem,''
	\textit{Computers \& Operations Research},
	vol. 184, Art. no. 107212, 2025.
	\href{https://doi.org/10.1016/j.cor.2025.107212}
	{doi: 10.1016/j.cor.2025.107212}.
	
	\bibitem{ref4}
	F. Wang, Y. Chen, L. He, J. Chen,
	S. Xu, and H. Huang,
	``A Lagrangian relaxation-based heuristic
	algorithm for multiple agile Earth observation
	satellite scheduling with time-dependent constraint,''
	\textit{European Journal of Operational Research},
	vol. 332, no. 1, pp. 84--100, 2026.
	\href{https://doi.org/10.1016/j.ejor.2025.12.047}
	{doi: 10.1016/j.ejor.2025.12.047}.
	
	\bibitem{ref5}
	M. Chen, B. Li, L. Xing, and Y. Chen,
	``DEEM: A differential evolution ensemble
	method for optimizing agile satellite scheduling,''
	\textit{Expert Systems with Applications},
	vol. 272, Art. no. 126771, 2025.
	\href{https://doi.org/10.1016/j.eswa.2025.126771}
	{doi: 10.1016/j.eswa.2025.126771}.
	
	\bibitem{ref6}
	B. Li, M. Chen, L. Xing, Y. Chen,
	and Y. Chen,
	``Optimizing time-dependent multi-agile
	Earth observation satellite scheduling problem
	using deep Q-learning and ensemble heuristics,''
	\textit{Information Sciences},
	vol. 712, Art. no. 122140, 2025.
	\href{https://doi.org/10.1016/j.ins.2025.122140}
	{doi: 10.1016/j.ins.2025.122140}.
	
	\bibitem{ref7}
	M. Chen, L. Wei, J. Chun, L. He,
	S. Xiang, L. Xing, and Y. Chen,
	``A neural priority model for agile Earth
	observation satellite scheduling using
	deep reinforcement learning,''
	\textit{Applied Soft Computing},
	vol. 174, Art. no. 112984, 2025.
	\href{https://doi.org/10.1016/j.asoc.2025.112984}
	{doi: 10.1016/j.asoc.2025.112984}.
	
	\bibitem{ref8}
	W. Huang, H. Wang, J. Wu, H. Hou,
	J. Li, Z. Li, and Y. Song,
	``A reinforcement learning-enhanced dung
	beetle optimization approach for agile
	Earth observation satellite scheduling,''
	\textit{IEEE Geoscience and Remote Sensing Letters},
	vol. 22, Art. no. 8000305, 2025.
	\href{https://doi.org/10.1109/LGRS.2025.3527925}
	{doi: 10.1109/LGRS.2025.3527925}.
	
	\bibitem{ref9}
	H. Wang, J. Wu, H. Li, Y. Song,
	W. Pedrycz, and L. Li,
	``Reinforcement learning-guided evolutionary
	policy optimization for preference-adjustable
	heterogeneous agile Earth observation satellite
	scheduling,''
	\textit{IEEE Transactions on Geoscience and
		Remote Sensing},
	pp. 1--16, 2026, early access.
	\href{https://doi.org/10.1109/TGRS.2026.3745218}
	{doi: 10.1109/TGRS.2026.3745218}.
	
	\bibitem{ref10}
	L. Wei, M. Chen, L. Xing, Q. Wan,
	Y. Song, Y. Chen, and Y. Chen,
	``Knowledge-transfer based genetic programming
	algorithm for multi-objective dynamic agile
	Earth observation satellite scheduling problem,''
	\textit{Swarm and Evolutionary Computation},
	vol. 85, Art. no. 101460, 2024.
	\href{https://doi.org/10.1016/j.swevo.2023.101460}
	{doi: 10.1016/j.swevo.2023.101460}.
	
	\bibitem{ref11}
	B. Romera-Paredes, M. Barekatain,
	A. Novikov, M. Balog, M. P. Kumar,
	E. Dupont, F. J. R. Ruiz, et al.,
	``Mathematical discoveries from program
	search with large language models,''
	\textit{Nature},
	vol. 625, no. 7995, pp. 468--475, 2024.
	\href{https://doi.org/10.1038/s41586-023-06924-6}
	{doi: 10.1038/s41586-023-06924-6}.
	
	\bibitem{ref12}
	N. van Stein and T. B\"ack,
	``LLaMEA: A large language model evolutionary
	algorithm for automatically generating metaheuristics,''
	\textit{IEEE Transactions on Evolutionary Computation},
	vol. 29, no. 2, pp. 331--345, 2025.
	\href{https://doi.org/10.1109/TEVC.2024.3497793}
	{doi: 10.1109/TEVC.2024.3497793}.
	
	\bibitem{ref13}
	M. Dorigo, V. Maniezzo, and A. Colorni,
	``Ant system: Optimization by a colony
	of cooperating agents,''
	\textit{IEEE Transactions on Systems,
		Man, and Cybernetics, Part B (Cybernetics)},
	vol. 26, no. 1, pp. 29--41, 1996.
	\href{https://doi.org/10.1109/3477.484436}
	{doi: 10.1109/3477.484436}.
	
	\bibitem{ref14}
	D. Karaboga and B. Basturk,
	``A powerful and efficient algorithm for
	numerical function optimization:
	Artificial bee colony (ABC) algorithm,''
	\textit{Journal of Global Optimization},
	vol. 39, no. 3, pp. 459--471, 2007.
	\href{https://doi.org/10.1007/s10898-007-9149-x}
	{doi: 10.1007/s10898-007-9149-x}.
	
	\bibitem{ref15}
	D. E. Goldberg,
	\textit{Genetic Algorithms in Search,
		Optimization, and Machine Learning}.
	Reading, MA, USA: Addison-Wesley, 1989.
	
	\bibitem{ref16}
	J. Kennedy and R. Eberhart,
	``Particle swarm optimization,''
	in \textit{Proceedings of the IEEE
		International Conference on Neural Networks},
	Perth, Australia, 1995,
	vol. 4, pp. 1942--1948.
	\href{https://doi.org/10.1109/ICNN.1995.488968}
	{doi: 10.1109/ICNN.1995.488968}.
	
	\bibitem{ref17}
	S. Mirjalili and A. Lewis,
	``The whale optimization algorithm,''
	\textit{Advances in Engineering Software},
	vol. 95, pp. 51--67, 2016.
	\href{https://doi.org/10.1016/j.advengsoft.2016.01.008}
	{doi: 10.1016/j.advengsoft.2016.01.008}.
	
\end{thebibliography}
\end{document}